%% file: main.tex
\ifdefined\XeTeXversion
  \AddToHook{package/microtype/after}{\renewcommand{\DisableLigatures}[2][]{}}
\fi

\documentclass[]{fairmeta}

\ifdefined\XeTeXversion
  \DeclareFontShape{T1}{optimistic}{m}{n}{<-> s * [0.88] "[assets/Optimistic.ttf]:mapping=tex-text"}{}
  \DeclareFontShape{T1}{optimistic}{b}{n}{<-> s * [0.88] "[assets/Optimistic.ttf]:mapping=tex-text"}{}
\fi

\usepackage{latexsym}
\usepackage[utf8]{inputenc}
\usepackage{xspace}
\usepackage{array}
\usepackage{amsmath}
\usepackage{amssymb}
\usepackage{enumitem}
\usepackage{tikz}
\usetikzlibrary{arrows.meta, positioning, calc}
\definecolor{tzblue}{HTML}{2A78D6}
\definecolor{tzorange}{HTML}{EB6834}
\definecolor{tzaqua}{HTML}{1BAF7A}
\definecolor{tzyellow}{HTML}{C98500}
\definecolor{tzmagenta}{HTML}{E87BA4}
\definecolor{tzgreen}{HTML}{1B7A1B}
\definecolor{tzgray}{HTML}{898781}
\definecolor{tzink}{HTML}{0B0B0B}

\newcommand{\sys}{\textsc{RAGCompass}\xspace}

\title{RAGCompass: Generalizable Retriever Routing via Reranking}

\author[1]{Valentin Six}
\author[2]{Mufei Li}
\author[3]{Dongqi Fu}
\author[3]{Yinglong Xia}
\author[3]{Hong Li}
\author[2]{Pan Li}

\affiliation[1]{University of Virginia}
\affiliation[2]{Georgia Institute of Technology}
\affiliation[3]{Meta}

\abstract{\input{sections/00_abstract}}

\date{\today}

\begin{document}

\maketitle

\section{Introduction}
\label{sec:introduction}
\input{sections/01_introduction}

\section{Related Work}
\label{sec:related_work}
\input{sections/02_related_work}

\section{Method}
\label{sec:method}
\input{sections/03_method}

\section{Experimental Setup}
\label{sec:experimental_setup}
\input{sections/04_experimental_setup}

\section{Results}
\label{sec:results}
\input{sections/05_results}

\section{Discussion}
\label{sec:discussion}
\input{sections/06_discussion}

\section{Conclusion}
\label{sec:conclusion}
\input{sections/08_conclusion}

\section*{Limitations}
\label{sec:limitations}
\input{sections/07_limitations}

\section*{Acknowledgements}
\label{sec:acknowledgements}
\input{sections/10_acknowledgements}

\clearpage
\newpage
\bibliographystyle{assets/plainnat}
\bibliography{references}

\clearpage
\newpage
\beginappendix
\label{sec:appendix}
\input{sections/09_appendix}

\end{document}

%% file: sections/00_abstract.tex
Different retrievers exist for retrieval-augmented generation, yet none performs best for all queries. Therefore, the key question is: given a query and retrieved passages from different retrievers, which one is most suited for that query? Existing routers usually make predictions from query-specific features and a fixed pool of retrievers; such routers have no output for an unseen retriever and are incapable by design on a new corpus. To fill these gaps, \sys reads the retrieved passage sets themselves: passages from an unfamiliar retriever or corpus stay directly comparable to other passages, so routing stays well designed where a query-only map would fail. We contrastively fine-tune a reranker against the answer each evidence set produces, which accounts for most of the gain; a small comparison head adds a moderate improvement, and neither component has retriever-specific parameters. \sys improves answer F1 by $28\%$ relative to the best fixed retriever, and outperforms fusion baselines, three query-only routers, and a prompted $8$B language model. Generalization experiments show that \sys keeps most of its gain under an unseen corpus, retriever, or reader, while query-only baselines fall to the best fixed retriever or below.

Code is made available at \url{https://github.com/Graph-COM/RAGCompass}.

%% file: sections/01_introduction.tex
\begin{figure}[t]
  \centering
  \includegraphics{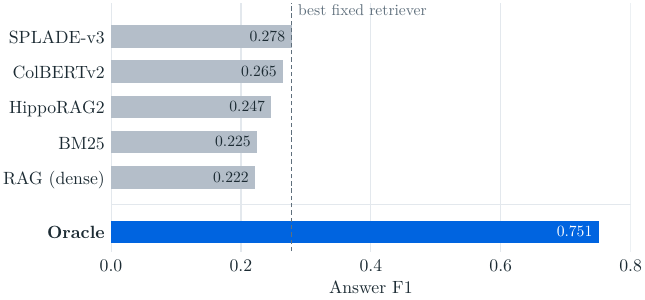}
  \caption{Answer quality for retrievers (grey), pooled across four QA datasets; comparison to an oracle (blue) selecting the best retriever per query.
  The retrievers are close to one another and all far below the oracle: a large room for improvement exists in choosing the best retriever per query at inference.}
  \label{fig:retriever-f1}
\end{figure}

In retrieval-augmented generation (RAG), a retriever first selects passages from a corpus, and a reader (e.g., a language model) then writes an answer from them.
RAG is widely used because it gives a language model access to knowledge beyond its training data and makes its answers traceable to their sources \citep{lewis_retrievalaugmented_2020}.
A deployed RAG system usually fixes one retriever for all queries, yet sparse lexical matching, learned sparse expansion, dense embeddings, late interaction, and graph-based linking \citep{robertson_probabilistic_2009,formal_splade_2021,lassance_spladev3_2024,karpukhin_dense_2020,lewis_retrievalaugmented_2020,santhanam_colbertv2_2022,gutierrez_hipporag_2024,gutierrez_rag_2025} each define a different notion of relevance for retrieval, and each one wins on different types of queries. Averaged over various datasets, the retrievers we study are close in performance ($0.22$ to $0.28$ answer F1), but an oracle that picks the right one for each query reaches $0.75$ F1, almost $3\times$ the best fixed retriever (Figure~\ref{fig:retriever-f1}). The opportunity is therefore not necessarily to design a better mechanism, but rather a way of choosing among existing ones given a specific query.

This is the problem of \emph{retriever routing}, posed as learning to rank retrievers by \citet{kim_ltrr_2025}: given a query and several structurally different retrievers, which retriever should one prefer? What makes routing hard in practice is that a deployed system changes: a text corpus evolves, retrievers are added or replaced, the reader is upgraded... Adapting the router after each  of those changes means re-running the reader over every training query to regenerate its labels, so a routing method is usable only if it holds on a corpus, mechanism pool, or reader it was not trained on. This practical necessity for routing robustness is left untouched by most prior work.

Indeed, most existing routers choose a retriever from query-specific features, without looking at any retrieved evidence: query performance prediction and learned routers predict which mechanism will do well from the query's wording, hand-designed features, or a retriever's own confidence scores \citep{meng_query_2023,chifu2026uncovering,tian_predicting_2026,lee2025routerretrieverroutingmixtureexpert,kalra_mor_2025,kim_ltrr_2025,zhao_r3ag_2026}. The resulting map is tied to a precise query and corpus distribution, and a specific pool of retrievers: a new corpus changes what is retrieved, and a new retriever is unknown to the system. To our knowledge, no previous closely related work focuses on a routing task under these changes.

We instead let the decision depend on the evidence coming from each retriever. Whichever mechanism or corpus produces the evidence, retrieved passages stay comparable to one another. Deciding which passages best answer the query should therefore be less sensitive to shifts in the query distribution, the retrievers, or the corpus. One disadvantage is that it requires running every candidate retriever first. We argue that the cost of retrieval is small, as the retrievals can run concurrently, and the decision adds about $0.2$s to a $5 - 6$s per query pipeline. To that end, we pose routing as \emph{selection among evidence sets}, which fixes two design choices: the scorer reads the passages themselves, so a \emph{reranker} is the natural building block; and it is trained on the answer the evidence yields rather than on document relevance, since the two often disagree and relevance-based training leads to lower answer quality (Section~\ref{sec:results}).

Our method, \sys, fine-tunes a reranker \citep{yang_qwen3_2025,zhang_qwen3_2025} so that its score for a mechanism's passages tracks the answer they produce, and adds a small comparison head that reads the reranker's representations of all candidates jointly. The fine-tuned reranker accounts for most of the accuracy gain; the head adds $1.4$pp on top. Neither the reranker nor the head holds retriever-specific parameters, so the router applies to a retriever it was not trained on.

Trained on four question-answering datasets and routing among five retrievers, \sys achieves a $28\%$ relative improvement over the best fixed retriever and beats a weighted mixture of retrievers, the strongest baseline we run. It also improves retrieval quality over the best fixed retriever ($+5.0$ points of recall@5, $+1.3$ of nDCG@5).

\sys also generalizes by design: held out from an entire dataset and its corpus, \sys retains more than $80\%$ of its in-distribution gain on some datasets (Figure~\ref{fig:crown}).

\paragraph{Contributions}
\begin{enumerate}[topsep=2pt,itemsep=2pt,leftmargin=*]
  \item \emph{Retriever routing as reranking over evidence:} we pose routing as choosing among the evidence sets the candidates returned, using the corresponding generated answers as the ranking criterion, instead of predicting a retriever's quality from simple query-specific signals.
  \item \sys \emph{outperforms existing routing approaches:} a reranker fine-tuned on answer utility beats rank-fusion and mixture baselines, three learned query-only routers, and an $8$B language model prompted to route, on answer F1.
  \item \emph{Routing generalization:} \sys carries to an unseen corpus, mechanism, and reader, while the three query-only routers do not.
\end{enumerate}

%% file: sections/02_related_work.tex
Methods for choosing among retrievers differ in what they condition on. We organize prior work around those differences; Table \ref{tab:salesman} summarizes.

\begin{table*}[t]
  \centering
  \footnotesize
  \setlength{\tabcolsep}{5pt}
  \begin{tabular}{@{}lcccccc@{}}
    \toprule
    \textbf{Property of the formulation} & \textbf{RRF} & \textbf{MoR} & \textbf{QPP\,/\,LTRR} & \textbf{R.RETRIEVER} & \textbf{R\textsuperscript{3}AG} & \textbf{\sys(ours)} \\
    \midrule
    Conditions on the query                  & $\times$ & \checkmark & \checkmark & \checkmark & \checkmark & \checkmark \\
    Reads the retrieved evidence             & $\times$ & $\times$ & $\times$   & $\times$   & $\times$ & \checkmark \\
    Supervised by answer utility             & $\times$ & $\times$ & \checkmark & \checkmark   & \checkmark & \checkmark \\
    Compares candidates jointly              & \checkmark & \checkmark & $\times$ & $\times$ & $\times$   & \checkmark \\
    Robust to an unseen corpus               & --- & --- & $\times$ & $\times$ & $\times$ & \checkmark \\
    Defined for an unseen mechanism          & \checkmark & \checkmark & $\times$ & $\times$ & \checkmark & \checkmark \\
    \bottomrule
  \end{tabular}
  \caption{\textbf{The gap we fill}: what each formulation reads, and whether it holds up under the two shifts we test (discussed above; measured in Section~\ref{sec:results-generalization}). ``---'': the shift cannot break a mapping the method never builds.}
  \label{tab:salesman}
\end{table*}

\paragraph{Routing that never reads the evidence} Reciprocal Rank Fusion \citep{cormack_reciprocal_2009} sums a rank-based score across systems without training: a nearly free baseline that is unable to identify the best retriever per query. MoR \citep{kalra_mor_2025} fuses heterogeneous retrievers through a weighted sum of their scores, with per-query, per-retriever weights computed zero-shot from the geometry of each retriever's embedding space; the static-weight variant we run as a baseline (Appendix~\ref{sec:appendix-r3ag}) is query-independent. Query performance prediction conditions on the query, predicting a retriever's effectiveness from query features or retrieval-score statistics \citep{meng_query_2023}; \citet{chifu2026uncovering} find its accuracy highly collection-dependent across the sparse, hybrid and dense families we route over, with marginal downstream gains, and \citet{tian_predicting_2026} extend it to RAG with QPP-style, reader-perplexity and document-quality features. Among learned routers, LTRR \citep{kim_ltrr_2025} learns to rank retrievers against downstream utility gain as we do, but reads hand-engineered query and score features rather than the passages. ROUTERRETRIEVER \citep{lee2025routerretrieverroutingmixtureexpert} routes over domain-specialized experts of a single architecture by training-free nearest-centroid similarity between the query and each expert's own embeddings, whereas we route across structurally different paradigms. 
All of these methods decide from signals other than the retrieved set whose quality is at issue.

\paragraph{Utility-supervised routing without evidence} R\textsuperscript{3}AG \citep{zhao_r3ag_2026} is closest to our method in spirit: an LLM's assessment of retrieved documents decides which retrievers count as training positives, but the router never reads those documents. Each retriever is a single learned identity token fused with the query through attention. The routing score, therefore, depends on the query and a fixed per-retriever profile, not on the retrieved evidence. A retriever added after training will get a token, so the method stays defined, but that new token never receives a gradient, and in our reimplementation the router does not select an unseen retriever more often where it is the right choice (Section~\ref{sec:results-generalization}). What separates it from our formulation is what is read at inference; we reimplement R\textsuperscript{3}AG, with the details listed in Appendix~\ref{sec:appendix-r3ag}, and test it under both shifts (Section~\ref{sec:results}).

To the best of our knowledge, none of these formulations have been tested under a shift. Rank fusion and MoR learn no mapping from queries to retrievers, so a corpus or mechanism change cannot break one. QPP and the learned routers do build one, but stays undefined for a retriever it never saw and untested on a new corpus. How routing behaves in such new conditions is the question that makes the paper's central test.

%% file: sections/03_method.tex
\begin{figure*}[t]
  \centering
  \begin{tikzpicture}[
      every node/.style={font=\small},
      box/.style={draw=black!65, line width=1.1pt, rounded corners=1.5pt, align=center, inner sep=2pt, minimum height=6mm},
      qbox/.style={box, fill=black!5, text width=12mm},
      bbbox/.style={box, fill=tzyellow!10, draw=tzyellow!85, text width=18mm},
      headbox/.style={box, fill=black!4, draw=black!80, text width=21mm},
      readerbox/.style={box, fill=tzgreen!8, draw=tzgreen!75, text width=12mm},
      chosenbox/.style={box, fill=tzgreen!12, draw=tzgreen!85, line width=1.4pt, text width=13mm},
      ansbox/.style={box, fill=tzgreen!12, draw=tzgreen!85, line width=1.4pt, text width=11mm},
      arr/.style={-{Latex[length=2mm,width=1.5mm]}, line width=1.1pt},
      tarr/.style={-{Latex[length=1.8mm,width=1.4mm]}, line width=0.9pt, dashed, draw=tzgray},
    ]

    \node[qbox] (q) at (-0.1,2.2) {Query\\$q$};

    \node[bbbox, minimum height=42mm] (bb) at (5.95,2.2)
      {Reranker\\backbone};
    \node[headbox, minimum height=42mm] (head) at (8.6,2.2)
      {Comparison\\head};


    \foreach \i/\col/\yy in {%
      1/tzblue/4.0,%
      2/tzorange/3.1,%
      3/tzaqua/2.2,%
      R/tzmagenta/0.4%
    } {
      \draw[line width=0.9pt, fill=\col!10, draw=\col!45] ($(3.9,\yy)+(-0.30,-0.34)$) rectangle ++(0.70,0.46);
      \draw[line width=0.9pt, fill=\col!16, draw=\col!60] ($(3.9,\yy)+(-0.25,-0.29)$) rectangle ++(0.70,0.46);

      \node[box, fill=\col!18, draw=\col!85, text width=19mm, font=\footnotesize, minimum height=7mm] (r\i) at (2.3,\yy) {Retriever $\i$};
      \node[box, fill=\col!24, draw=\col!95, text width=6mm] (d\i) at (4.1,\yy) {$D_\i$};

      \draw[arr, draw=\col!85] (q.east) to[out=0,in=180] (r\i.west);
      \draw[arr, draw=\col!85] (r\i.east) -- (d\i.west);
      \draw[arr, draw=\col!85] (d\i.east) -- (bb.west |- d\i.east);
      \draw[arr, draw=\col!70] (bb.east |- d\i.east) -- (head.west |- d\i.east);
    }
    \node[font=\Large, tzink] at (2.3,1.4) {$\vdots$};
    \node[font=\Large, tzink] at (4,1.4) {$\vdots$};
    \node[font=\small, tzink] at (7.20,4.40) {$h_i$};

    \node[chosenbox] (chosen) at (10.95,2.2) {Chosen\\$D_{\hat r}$};
    \node[readerbox] (reader) at (12.85,2.2) {Reader};

    \node[ansbox] (ans) at (14.50,2.2) {Answer};
    \draw[tzgreen, line width=1.1pt] (14.50,2.85) circle (0.13);
    \draw[tzgreen, line width=1.1pt] (14.50-0.06,2.85-0.01) -- (14.50-0.01,2.85-0.07) -- (14.50+0.08,2.85+0.07);

    \draw[arr, draw=tzgreen, line width=1.4pt] (head.east) -- (chosen.west);
    \draw[arr, draw=tzgreen] (chosen.east) -- (reader.west);
    \draw[arr, draw=tzgreen] (reader.east) -- (ans.west);

    \node[draw=tzgray, dashed, line width=0.9pt, rounded corners=2pt,
          fill=black!2, align=center, inner sep=4pt, text width=132mm]
      (sup) at (7.6,-1.05)
      {\footnotesize \emph{Training only}: the reader answers from \emph{every} candidate's evidence, giving $y_i=\mathrm{F1}(A(q,D_i),a^\star)$;\\
       the scores are fit to that utility rather than to retrieval relevance, by $\mathcal{L}=\mathrm{KL}(u\,\|\,p)$.};
    \draw[tarr] (5.9,-0.52) -- (5.9,0.10);
    \draw[tarr] (8.6,-0.52) -- (8.6,0.10);
    \draw[tarr] (12.9,1.85) -- (12.9,-0.52);

  \end{tikzpicture}
  \caption{\textbf{The formulation, end to end}: each candidate mechanism retrieves an evidence set $D_i$; a shared reranker backbone scores each $(q,D_i)$ pair in its own forward pass, reading the passages rather than the mechanism's identity or confidence score; a comparison head with no mechanism-specific parameters reads $h_1,\dots,h_R$ jointly; and the reader answers from the selected set $D_{\hat r}$. Supervision is the answer each evidence set yields (dashed, training only). Nothing in the pipeline changes if a mechanism is added, removed, or run on a different corpus.}
  \label{fig:teaser}
\end{figure*}
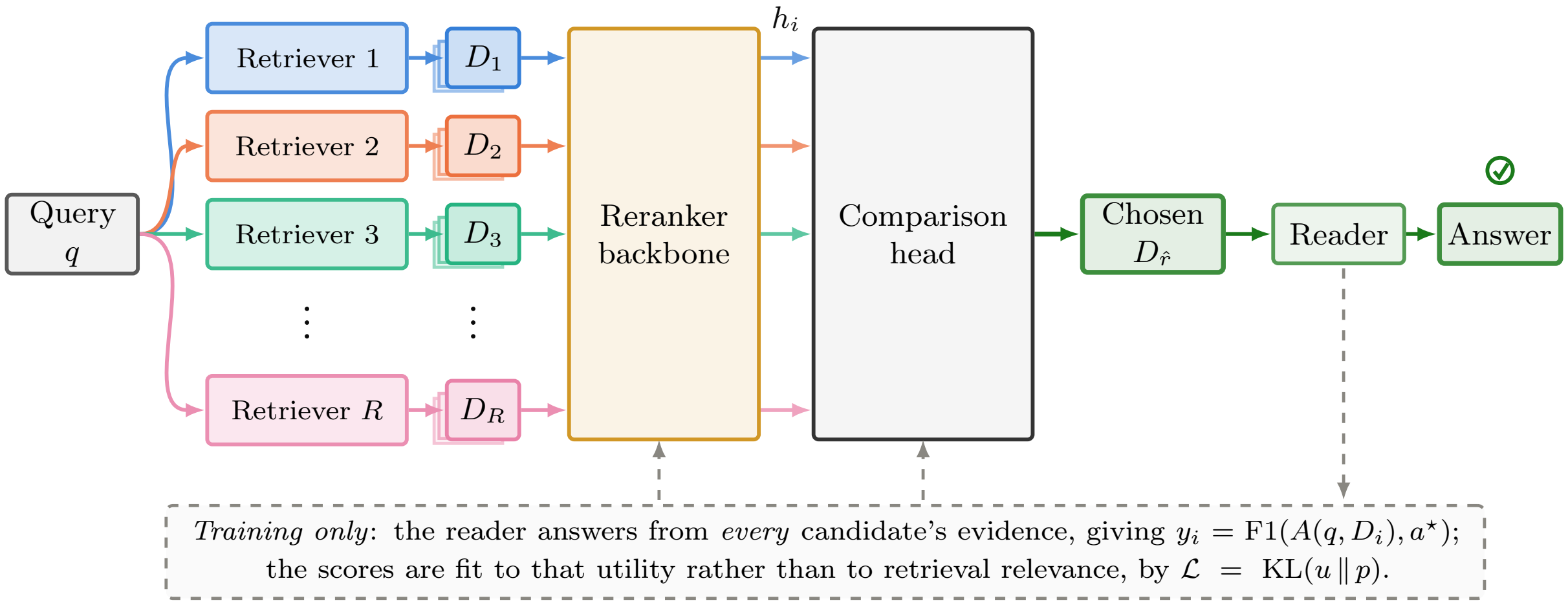

\subsection{Problem formulation}
\label{sec:formulation}

Let a query $q$ be served by one of $R$ retrievers, where retriever $i$ returns a passage set $D_i = (d_{i_1}, \dots, d_{i_k})$, and let a fixed reader $A$ produce an answer $A(q, D_i)$ from whichever set it is given. We define the \emph{utility} of retriever $i$ on $q$ as the quality of the answer that its evidence actually yields,
\begin{equation}
y_i = \mathrm{F1}\big(A(q, D_i),\, a^\star\big), \quad i = 1, \dots, R,
\label{eq:utility}
\end{equation}
for gold answer $a^\star$. Routing seeks $\hat r$ to maximize $y_{\hat r}$, given $q$ and $D_1, \dots, D_R$ but not $a^\star$. This is the reranking problem announced in Section~\ref{sec:introduction}: the candidates to be ranked are the evidence sets $D_1, \dots, D_R$, and the ranking criterion is $y_i$.

Two properties of Equation~\ref{eq:utility} guide our method design. First, \emph{content grounding}: $y_i$ is a property of $D_i$, so a router that never reads $D_i$ predicts $y_i$ from a proxy. Second, \emph{utility supervision}: $y_i$ is defined through the reader, not through retrieval relevance, since a passage set can be topically relevant and still yield a wrong answer. The two diverge in practice: nDCG against gold passages and answer F1 select different retrievers on roughly $43\%$ of queries,
so a router trained on relevance optimizes toward the wrong retriever on close to half the queries.

We accordingly train a scorer $s_i = f_\theta(q, D_i)$ against the utility vector, converting both to distributions over the candidate set and using KL divergence \cite{kullback1951information}:
\begin{equation}
p_i = \frac{\exp(s_i)}{\sum_j \exp(s_j)}, \qquad
u_i = \frac{\exp(y_i / \tau)}{\sum_j \exp(y_j / \tau)},
\label{eq:softtarget}
\end{equation}
\begin{equation}
\mathcal{L} = \mathrm{KL}(u \,\|\, p).
\label{eq:kl}
\end{equation}
The shared normalization in Equation~\ref{eq:softtarget} makes the gradient on $s_i$ depend on all other $s_j$, creating competition for probability mass consistent with selecting one retriever per query. Section~\ref{sec:results-effective} measures what this coupling is worth. The temperature $\tau$ sets how strictly the target favors the arg max. Since $f_\theta$ must read passage text, a reranker that models query--passage pairs is a natural backbone.

\subsection{Pipeline}

\paragraph{Per-retriever encoding and scoring} We encode each retriever's passages in its own forward pass through a shared backbone, Qwen3-Reranker-0.6B \citep{yang_qwen3_2025,zhang_qwen3_2025}, and read a scalar score from the logit of the token \texttt{yes} under a routing-oriented instruction:
\begin{equation}
s_i = \mathrm{logit}_\theta(\texttt{yes} \mid q, D_i).
\label{eq:yeslogit}
\end{equation}
The alternative is to concatenate all $R$ retrievers' passages into one prompt and read off per-retriever positions. This would let the backbone itself compare the candidates, but it underperforms per-retriever encoding in every variant we tried (Appendix~\ref{sec:appendix-joint}), as the per-retriever passage budget shrinks with $R$. Fine-tuned on its own, this scorer is our first model (``scalar scorer'' in Table~\ref{tab:main}).

\paragraph{Comparison head} Scoring alone reduces each candidate to a single number before any comparison takes place; that means the ranking is done over $R$ scalars rather than over the evidence that Equation~\ref{eq:utility} says the decision depends on. To compare representations directly instead, we take the backbone's per-retriever hidden states $h_i$ and pass all $R$ jointly through a two-layer transformer encoder \citep{vaswani2017attention} whose self-attention runs across the retrievers, followed by a linear scorer: 
\begin{align}
z_1, \dots, z_R &= \mathrm{Head}(h_1, \dots, h_R), \\
\hat r &= \arg\max_i z_i,
\label{eq:head}
\end{align}
trained with the same objective (Equations~\ref{eq:softtarget}--\ref{eq:kl}) over $z$. Similar to the scorer, the head has no parameters tied to particular retrievers, so the full router stays valid for a candidate pool it was not trained on. A variant that adds a learned per-retriever identity embedding to $h_i$ loses this property (Section~\ref{sec:results}). The head costs under $1\%$ of the routing decision's time; Section~\ref{sec:results-effective} reports its contribution to accuracy. We consider one pooled vector per retriever or one hidden state per passage for $h_i$.

\paragraph{Training the head} We first fine-tune the backbone alone against Equation~\ref{eq:kl}, then train the head on its frozen representations. This two-stage recipe matches joint training of head and backbone at three seeds (Section~\ref{sec:results}), and it is the configuration behind every generalization result.

%% file: sections/04_experimental_setup.tex
\paragraph{Datasets and retrievers} We jointly train and evaluate on four single-hop and multi-hop QA datasets: Natural Questions \citep{kwiatkowski_natural_2019}, MuSiQue \citep{trivedi_musique_2022}, HotpotQA \citep{yang_hotpotqa_2018}, and 2WikiMultihopQA \citep{ho_constructing_2020}. Each query is routed among $R{=}5$ retrievers from deliberately different paradigms, which is what makes their confidence scores incommensurable and distinguishes our setting from routing among same-architecture experts \citep{lee2025routerretrieverroutingmixtureexpert}: a dense retriever \citep{karpukhin_dense_2020,lewis_retrievalaugmented_2020}, BM25 \citep{robertson_probabilistic_2009}, SPLADE-v3 \citep{formal_splade_2021,lassance_spladev3_2024}, ColBERTv2 \citep{santhanam_colbertv2_2022}, and HippoRAG2 \citep{gutierrez_hipporag_2024,gutierrez_rag_2025}. Each supplies $k{=}5$ passages to a fixed Llama-3.1-8B-Instruct \cite{grattafiori2024llama3herdmodels} reader. Queries on which all retrievers reach near-identical utility give a nearly uniform target $u$, carrying no routing signal but adding gradient noise.

\paragraph{Metric and splits} We report test-set answer F1 from the selected retriever's passages; this corresponds to the utility of Equation~\ref{eq:utility}, consistent with the RPP/GPP framing of \citet{tian_predicting_2026}. We additionally report the recall@5 and nDCG@5 scores of the routed choice. We use a 70/30 train/validation resplit. We train and evaluate only on queries with a top-two utility gap $\geq 0.05$.
We average the main result, both ablations, and generalization protocols over three seeds.

\paragraph{Baselines} We report the \textbf{Oracle} (best retriever per query) as an upper-bound for router performance, and the \textbf{best fixed retriever}. Non-learned combinations are \textbf{RRF}~\citep{cormack_reciprocal_2009} (unsupervised rank fusion, $k{=}60$) and a \textbf{MoR}-style mixture~\citep{kalra_mor_2025} (static retriever weights fit on validation utility). Learned routers condition on the query but never read passages when routing; all three are our reimplementations, adapted to our retriever pool and compute budget, and Appendix~\ref{sec:appendix-r3ag} lists how each differs from its original: the \textbf{LTRR-style ranker}~\citep{kim_ltrr_2025} (query and retrieval-confidence features, trained on our utility targets, model family chosen on validation F1), \textbf{R.RETRIEVER}~\citep{lee2025routerretrieverroutingmixtureexpert} (training-free nearest-centroid routing on training queries where each retriever's F1 is highest), and \textbf{R\textsuperscript{3}AG}~\citep{zhao_r3ag_2026} (two contrastively trained Qwen3-Embedding-0.6B encoders with one learned identity token per retriever). Two answer-level baselines complete the set: \textbf{majority vote} over the $R$ generated answers, and \textbf{LLM-as-router}, where Llama-3.1-8B-Instruct picks zero-shot among the $R$ evidence sets or answers.

\paragraph{Generalization protocols} We test three shifts, each changing one component of the pipeline while the rest stays fixed. \emph{Unseen corpus}: leave-one-dataset-out trains on three datasets and evaluates on the fourth, whose queries and passages are never seen in training. This gives one fold per dataset; the control trains on all four and is evaluated on the same rows, so the difference between the two isolates the cost of the shift. \emph{Unseen mechanism}: leave-one-retriever-out removes one retriever from the training pool, together with its utility labels, but keeps it among the candidates at test time. This gives one fold per retriever; the control trains with all five. In both protocols the backbone and the head are retrained for each fold, so neither component has seen the hidden dataset or retriever, and the compared baselines are run under the identical protocol. \emph{Unseen reader}: the router, trained on utility labels from Llama-3.1-8B-Instruct \cite{grattafiori2024llama3herdmodels}, and its evidence are left unchanged, and Qwen2.5-7B-Instruct \cite{qwen2025qwen25technicalreport} answers in place of Llama-3.1-8B-Instruct. These three shifts mirror how a deployed RAG system changes over time: the corpus evolves, retrievers are added or replaced, and the reader is upgraded. Since collecting utility labels requires the reader to answer from every candidate's evidence, retraining the router after each such change is costly, and a router is only practical if its gains survive them.

%% file: sections/05_results.tex

\subsection{\sys outperforms routing baselines, at a small added cost}
\label{sec:results-effective}

\paragraph{Answer quality} Table~\ref{tab:main} groups methods by input. Query-independent methods (RRF and MoR) beat the best fixed retriever but plateau at $0.32$. Three learned query-only routers are query-conditioned and trained against answer utility, yet fall below both: the LTRR-style ranker reaches performance within noise of the best fixed retriever; R\textsuperscript{3}AG underperforms it; and R.RETRIEVER achieves the lowest performance of the three. Our scalar scorer, identically supervised but reading the passages, reaches $0.3412$, and the comparison head brings \sys to $0.3552$: $+7.7$pp over the best fixed retriever and $+3.5$pp over MoR (best measured baseline). Most of this gain comes from the fine-tuned scorer ($6.3$ of the $7.7$pp). The head adds $1.4$pp, a small improvement against seed standard deviations of $0.3$--$0.6$pp; we keep it because the gain exceeds that variation and because the head, like the scorer, has no per-retriever parameters, so adding it does not tie the router to the retrievers seen in training (Section~\ref{sec:results-generalization}).

\begin{table}[h]
  \centering
  \small
  \begin{tabular}{@{}lr@{\,}l@{}}
    \toprule
    \textbf{Method} & \multicolumn{2}{c}{\textbf{Test F1}} \\
    \midrule
    Best fixed retriever & 0.2782 \\
    \addlinespace
    \multicolumn{3}{@{}l}{\emph{Not query-conditioned}} \\
    \quad RRF & 0.3002 \\
    \quad MoR (static weights) & 0.3203 \\ 
    \addlinespace
    \multicolumn{3}{@{}l}{\emph{Query only}} \\
    \quad LTRR-style ranker & 0.2795 & {\scriptsize $\pm$0.0019} \\
    \quad R\textsuperscript{3}AG & 0.2649 & {\scriptsize $\pm$0.0012} \\
    \quad R.RETRIEVER & 0.2608 & {\scriptsize $\pm$0.0034} \\
    \addlinespace
    \multicolumn{3}{@{}l}{\emph{RAGCompass (ours)}} \\
    \quad Scalar scorer, no head & 0.3412 & {\scriptsize $\pm$0.0059} \\
    \quad \textbf{+ comparison head} & \textbf{0.3552} & {\scriptsize $\pm$0.0026} \\
    \midrule
    Oracle & 0.7508 \\
    \bottomrule
  \end{tabular}
  \caption{\textbf{Main result}: $+7.7$pp over best fixed, $+3.5$pp over MoR (strongest baseline). Trained methods: mean $\pm$ std over three seeds. Best fixed, RRF, MoR and the oracle are deterministic, so they have no seed variance; MoR uses static weights (Appendix~\ref{sec:appendix-r3ag}). Scored on the routable population (top-two utility gap $\geq 0.05$); the gain over unfiltered queries is reported in Section~\ref{sec:results-filter}.}
  \label{tab:main}
\end{table}

\paragraph{Why query-only routers fail to beat a fixed retriever} A router can score below the best fixed retriever only if it routes many queries to a worse one: the query-only routers concentrate their choices on very few retrievers instead of discriminating between queries. R\textsuperscript3AG is the clearest case: across seeds, it routes nearly all test queries to ColBERTv2, while SPLADE-v3 leads on test. No configuration in its hyperparameter grid beats the best fixed retriever on validation, and configurations that spread their choices more evenly score lower still (Appendix~\ref{sec:appendix-r3ag}). The LTRR-style ranker and R.RETRIEVER show the same tendency.

One likely reason for this behaviour, which we cannot fully isolate, is that the setting gives a query-only router little to learn from. We pool single-hop and multi-hop datasets; the five retrievers are close on average ($0.22$--$0.28$ F1), and the best one changes across datasets and even between the validation and test splits. When the query alone does not predict which retriever wins, the lowest-loss solution is to prefer the retriever that wins most often in training, and small differences between splits then decide whether that choice lands on the best fixed retriever or under it. The original methods were evaluated in settings where the query is more informative about the right source, for example routing between domain-specialized experts.

The routed choice also raises recall@5 by $5.0$ points and nDCG@5 by $1.3$ over the best fixed retriever (Appendix~\ref{sec:appendix-retrieval}).

\paragraph{Source of the gain: supervision, not joint normalization\\}

\emph{1) Content grounding}: the LTRR-style ranker and our scalar scorer are both query-conditioned and utility-supervised, and differ in what they read, but also in model class and capacity (a feature-based classifier against a fine-tuned $0.6$B reranker). The gap between them is therefore an upper bound on what reading the evidence is worth.

\emph{2) Utility supervision}: holding the example set, head and evaluation fixed and changing only the target, answer utility reaches $0.3552 \pm 0.0026$ versus $0.3341 \pm 0.0010$ for gold-passage nDCG. The gain does not rest on how the candidates' scores are trained. Coupling them through the shared normalization of Equation~\ref{eq:softtarget} is worth $0.54$pp on a frozen head, and end-to-end pointwise training reaches $0.3557 \pm 0.0047$ against $0.3522 \pm 0.0068$ for the listwise control. Both gaps fall within the seed spread that sets the resolution of every comparison here, so the objective can be chosen for simplicity without cost to routing quality.

\paragraph{Per-dataset gains and cost} The gain is uneven across datasets: roughly $12$pp on HotpotQA and 2WikiMultihopQA, $4.8$pp on MuSiQue, and $0.42$pp on NQ; NQ is also the dataset with the least room between its best fixed retriever and its oracle (Appendix~\ref{sec:appendix-perdataset}). A larger model does not close it: Llama-3.1-8B-Instruct prompted for zero-shot routing, about $13\times$ our router's parameters, gains at most $3.20$pp over the best fixed retriever. Our routing decision adds $0.222$s per query to a $5.687$s retrieve-and-answer pipeline, $3.9\%$ (Appendix~\ref{sec:appendix-cost}).

\subsection{The filter and the remaining oracle gap}
\label{sec:results-filter}

\begin{figure}[t]
  \centering
  \includegraphics{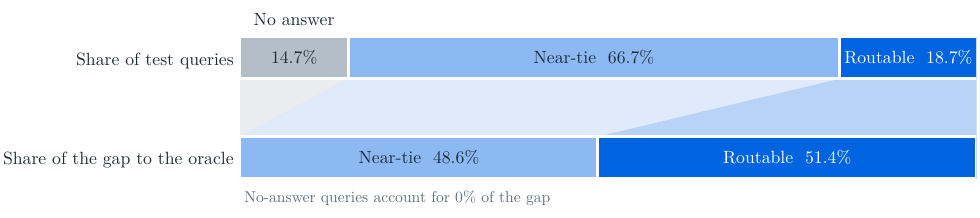}
  \caption{\textbf{The reported population holds half of the gap to the oracle}: the unfiltered test set split into three groups, by share of queries (top) and by share of the F1 gap between our router and the oracle (bottom). \emph{No answer}: no retriever's evidence yields a correct answer, so these queries add nothing to the gap. \emph{Near-tie}: the two best retrievers are within $0.05$ F1 of each other. \emph{Routable}: the two best retrievers differ by at least $0.05$; this is the population we report on, $18.7\%$ of queries holding $51.4\%$ of the gap.}
  \label{fig:oracle-buckets}
\end{figure}

\paragraph{Effect of the filter} All results above are scored on the routable population: queries whose top-two utility gap is at least $0.05$. Figure~\ref{fig:oracle-buckets} shows what this leaves out: on the unfiltered test set, no retriever answers $14.7\%$ of queries, and a further $66.7\%$ are near-ties between the two best retrievers; the remaining $18.7\%$ is the population we report on. Over every query, the gain shrinks from $+7.70$pp to $+2.28$pp. The best fixed retriever also changes (SPLADE-v3 to HippoRAG2), so the two gains differ in reference point as well as in population.

\paragraph{The oracle gap} A large gap to the oracle remains ($0.7508$ against $0.3552$), and on the reported population all of it is contested by construction. The router is not degenerate there: it matches the oracle on $32.2\%$ of queries, and $63.2\%$ of its misses still beat the best fixed retriever.

\subsection{\sys generalizes; query-only routers do not}
\label{sec:results-generalization}

\begin{table}[t]
  \centering
  \small
  \setlength{\tabcolsep}{3pt}
  \begin{tabular}{@{}lrrrc@{}}
    \toprule
    \textbf{Dataset} & \textbf{In-dist.} & \textbf{Held out} & \textbf{Fixed} & \textbf{Ret.} \\
    \midrule
    HotpotQA & 0.4168 {\scriptsize $\pm$0.0017} & 0.4210 {\scriptsize $\pm$0.0010} & 0.2928 & 103\% \\
    2Wiki & 0.3614 {\scriptsize $\pm$0.0020} & 0.3416 {\scriptsize $\pm$0.0082} & 0.2393 & 84\% \\
    MuSiQue & 0.3178 {\scriptsize $\pm$0.0042} & 0.2935 {\scriptsize $\pm$0.0044} & 0.2700 & 49\% \\
    NQ & 0.3495 {\scriptsize $\pm$0.0058} & 0.3478 {\scriptsize $\pm$0.0033} & 0.3453 & -- \\
    \bottomrule
  \end{tabular}
  \caption{\textbf{Survives an unseen corpus}: the router trained without a dataset's queries or corpus (``Held out'') against the same router trained with them (``In-dist.''), on the same rows. ``Fixed'': that dataset's best fixed retriever; ``Ret.'': held-out gain over best fixed as a share of in-distribution gain. Results averaged over three seeds.}
  \label{tab:corpus-transfer}
\end{table}

All preceding results are in-distribution. We now test four shifts; for unseen corpus and mechanism we also run the LTRR-style ranker and R\textsuperscript{3}AG under the identical protocol, and for the unseen reader and retrieval depth we report our own router only.

\begin{figure}[t]
  \begin{minipage}[t]{0.485\textwidth}
    \centering
    \includegraphics{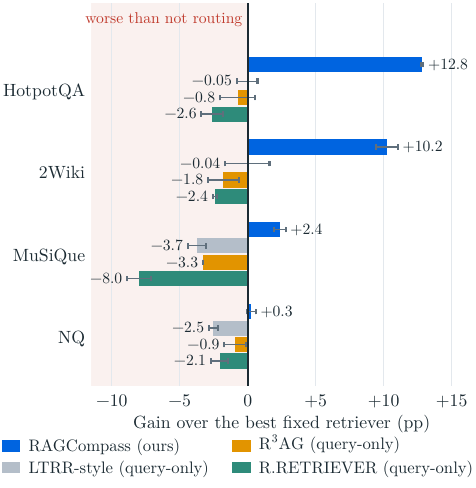}
    \caption{\textbf{Generalizable routing}: gain over each dataset's best fixed retriever when that dataset (its queries \emph{and} its corpus) is held out of training entirely. \sys stays positive on all four; three query-only routers, held to the identical protocol on the identical rows, never exceed the best fixed retriever. Three seeds for every method; whiskers are one std.}
    \label{fig:crown}
  \end{minipage}\hfill
  \begin{minipage}[t]{0.485\textwidth}
    \centering
    \includegraphics{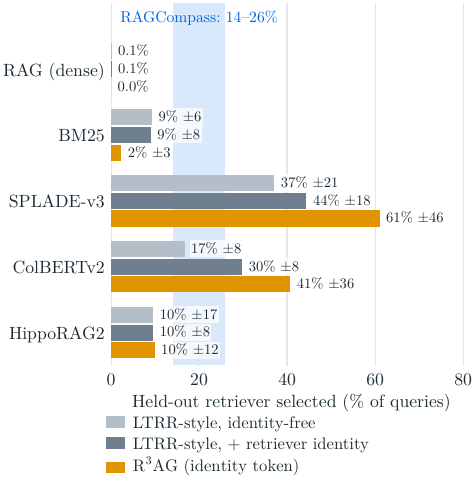}
    \caption{\textbf{Query-only routers do not discriminate}: how often each router selects the retriever held out of its training (three seeds, $\pm$ one std in each label). Shaded: the $14$--$26\%$ range of our head. On every fold the rate is within about three points of the rate on queries where the held-out retriever is the best choice.}
    \label{fig:ltrr-mechanism}
  \end{minipage}
\end{figure}

\paragraph{Unseen corpus} Holding out either of the two datasets with the largest in-distribution gains costs little (Table~\ref{tab:corpus-transfer}): HotpotQA retains its full in-distribution gain, 2WikiMultihopQA $84\%$, and MuSiQue about half. NQ's negligible gain ($+0.42$pp) makes its transfer result uninformative, so we do not average across datasets. Appendix~\ref{sec:appendix-datasize} shows that training on a quarter less data costs little by itself. All three query-only routers fail to improve (Figure~\ref{fig:crown}). The LTRR-style ranker falls slightly below the best fixed retriever on both MuSiQue ($-3.73$pp) and NQ ($-2.54$pp) and is level on HotpotQA and 2WikiMultihopQA. R\textsuperscript{3}AG is below on all four ($-0.75$ to $-3.30$pp), and so is R.RETRIEVER ($-2.07$ to $-8.01$pp). Our router gains $+10$ to $+13$pp on the two datasets with the most headroom.

\begin{table}[t]
  \centering
  \small
  \setlength{\tabcolsep}{4pt}
  \begin{tabular}{@{}lrrr@{}}
    \toprule
    \textbf{Held out} & \textbf{Test F1} & \textbf{Cost} & \textbf{Oracle pick} \\
    \midrule
    \multicolumn{4}{@{}l}{\emph{Identity-free head (three seeds)}} \\
    RAG (dense) & $0.3435 \pm 0.0049$ & $-1.17$pp & 19.7\% \\
    BM25 & $0.3494 \pm 0.0052$ & $-0.58$pp & 24.4\% \\
    SPLADE-v3 & $0.3485 \pm 0.0061$ & $-0.67$pp & 39.7\% \\
    ColBERTv2 & $0.3479 \pm 0.0062$ & $-0.73$pp & 33.0\% \\
    HippoRAG2 & $0.3536 \pm 0.0021$ & $-0.16$pp & 33.8\% \\
    \multicolumn{4}{@{}l}{\emph{+ identity embedding}} \\
    RAG (dense) & $0.3447 \pm 0.0032$ & $-1.05$pp & 1.1\% \\
    BM25 & $0.3383 \pm 0.0041$ & $-1.69$pp & 0.4\% \\
    SPLADE-v3 & $0.3285 \pm 0.0054$ & $-2.67$pp & 83.4\% \\
    ColBERTv2 & $\mathbf{0.2706} \pm 0.0031$ & $\mathbf{-8.46}$\textbf{pp} & 99.8\% \\
    HippoRAG2 & $0.3277 \pm 0.0048$ & $-2.75$pp & 10.7\% \\
    \bottomrule
  \end{tabular}
  \caption{\textbf{Survives an unseen mechanism}: leave-one-retriever-out, one fold per row; control with all five retrievers: $0.3552 \pm 0.0026$. ``Oracle pick'': held-out retriever selection rate when optimal.}
  \label{tab:mechanism-transfer}
\end{table}

\paragraph{Unseen mechanism} Our identity-free head costs $0.16$--$1.17$pp when a retriever is held out of training (Table~\ref{tab:mechanism-transfer}) and still selects that retriever $14$--$26\%$ of the time, and on $20$--$40\%$ of the queries where it is the oracle choice. The backbone and the head are both retrained without the held-out retriever, so no part of the pipeline has seen its passages or labels.
Naturally, adding a learned identity embedding makes the head degenerate on the retriever it never saw during training: it almost never picks BM25 or RAG ($0.1$--$0.2\%$), nearly always picks ColBERTv2 or SPLADE-v3 ($75$--$99\%$), and its ColBERTv2 fold falls to $0.2706$, below the best fixed retriever ($0.2782$). An $R$-way classifier cannot express this protocol, since it never predicts a class absent from training, so we re-pose the LTRR-style ranker as a \emph{shared} scorer over (query, retriever) pairs with no per-retriever parameters: it gains nothing in distribution ($-0.51 \pm 0.80$pp) and is below the best fixed retriever on all five folds; with a retriever-identity feature it is level in distribution and below on every fold. R\textsuperscript{3}AG, whose token for a new retriever is defined but untrained, is also below on every fold. None of these baselines discriminates between queries (Figure~\ref{fig:ltrr-mechanism}): their pick rates for the held-out retriever range from $0\%$ to $61\%$ across folds and stay nearly unchanged on the queries where it is the oracle choice, whereas our head picks it more often where it is right. 

\paragraph{Unseen reader} Leaving the router and its evidence untouched and re-answering with a new reader (Qwen2.5-7B-Instruct in place of Llama-3.1-8B-Instruct), the router retains $71\% \pm 2\%$ of its advantage; the best fixed retriever's identity changes with the reader (SPLADE-v3 to HippoRAG2), so this is a different reference point as well as a different reader. A fourth shift points the same way: evaluated at retrieval depth $k{=}3$ rather than $k{=}5$ (training was done at that $k=5$ setting) with no retraining, the router still beats that depth's best fixed retriever by $+3.64$pp and captures a comparable share of a smaller headroom (Appendix~\ref{sec:appendix-depth}).

%% file: sections/06_discussion.tex
Our experiments show that routing signals matter more than score coupling: utility supervision beats relevance supervision by $2.11$pp, while coupling the candidates' scores through joint normalization moves the result by less than seed noise (Section~\ref{sec:results-effective}). This sharpens our disagreement with R\textsuperscript{3}AG \citep{zhao_r3ag_2026}: the substantive difference is its representation of each retriever as a fixed identity token, where we use the evidence retrieved for the particular query; its per-retriever scoring is not the issue. Comparison design also matters less for accuracy than expected: comparing representations rather than scalars adds little (Section~\ref{sec:results-effective}), and
reading all retrievers in one context underperforms per-retriever encoding (Appendix~\ref{sec:appendix-joint}).

Reading the evidence is also what allows generalization: a passage set stays comparable to its rivals whatever produced it, whereas a map from query-side features or retriever identity has no output for a mechanism it was not fitted to. Consistent with this, none of the tested query-only routers exceeds the best fixed retriever under either shift, and our own head degrades once it is given a per-retriever identity embedding (Section~\ref{sec:results-generalization}). The query-only routers do not exceed the best fixed retriever in distribution either (Section~\ref{sec:results-effective}), so the comparison supports the claim that a query-only signal is weak in our pooled setting more than a claim about those systems as originally published.

A large oracle gap remains ($0.7508$ against $0.3552$). Section~\ref{sec:results-filter} shows: the router matches the oracle on about a third of the queries, and most of its misses still beat the best fixed retriever.

%% file: sections/08_conclusion.tex
In this work we study retriever routing: given a query and the passage sets returned by several retrievers, deciding which set the reader should see. We pose this retriever-routing problem as a content-grounded comparison supervised by downstream utility, rather than as predicting each retriever's quality from other signals. \sys uses a fine-tuned reranker as the scorer and a comparison head, neither with parameters tied to any retriever. \sys outperforms every router we compare against on answer F1, at $3.9\%$ of the cost of the generation pipeline; most of that gain comes from the utility-supervised reranker, and the head adds a small further improvement. More importantly, \sys keeps most of its advantage on a corpus, a retriever, or a reader it never saw in training, while the query-only routers do not exceed the best fixed retriever.
 Narrowing the remaining gap to the oracle is left for future work.

%% file: sections/07_limitations.tex
Utility (Equation~\ref{eq:utility}) is defined through one reader (Llama-3.1-8B-Instruct) and one retrieval depth ($k{=}5$, fixed at retrieval time), and the unseen-reader protocol tests exactly one alternative (Qwen2.5-7B-Instruct), so we can neither separate evidence quality from evidence--reader fit nor claim the reader result holds beyond that single swap. Depth is tested in one direction only: the stored retrievals are five deep and each row's utility is the answer the reader gave from a five-passage context, so going deeper would mean re-running retrieval for every retriever, dataset and split and regenerating every label -- how routing behaves as retrieval \emph{deepens} is untested. Every result uses one backbone scale (Qwen3-Reranker-0.6B), so whether the properties of Section~\ref{sec:formulation} hold at a different parameter count is unmeasured.

The three learned query-only baselines (LTRR, R\textsuperscript{3}AG \citep{zhao_r3ag_2026} and ROUTERRETRIEVER \citep{lee2025routerretrieverroutingmixtureexpert}) are our reimplementations, and they may differ from the original implementation by the authors; Appendix~\ref{sec:appendix-r3ag} lists the differences. Our MoR baseline is likewise a simplified variant: it uses one validation-fitted weight per retriever in place of the original's per-query weights.
The $3.9\%$ inference overhead covers the routing \emph{decision} only, on one device (L40S) at one batch size: routing also requires running all $R$ retrievers rather than one, a real cost that is parallelizable but not included in that figure, which should be read as an order of magnitude rather than a portable constant.

%% file: sections/10_acknowledgements.tex
Valentin Six, Mufei Li, and Pan Li are partially supported by the NSF under awards IIS-2239565, CCF-2402816, IIS-2435957; the Meta Grant; the NVIDIA Academic Grant Program; and the IDEaS Cyberinfrastructure Awards.

%% file: sections/09_appendix.tex
\section{Joint-prompt Encoding Ablation}
\label{sec:appendix-joint}

\begin{table}[h]
  \centering
  \small
  \setlength{\tabcolsep}{3pt}
  \begin{tabular}{@{}p{0.28\linewidth}p{0.42\linewidth}r@{\,}l@{}}
    \toprule
    \textbf{Variant} & \textbf{Extraction} & \multicolumn{2}{c}{\textbf{Test F1}} \\
    \midrule
    Joint, end-to-end (15 ep) & Section-end yes-logit per retriever & 0.2942 & {\scriptsize $\pm$0.0048} \\
    Joint, frozen (20 ep) & Final-token hidden state, linear head & 0.2735 & {\scriptsize $\pm$0.0031} \\
    Joint, end-to-end (10 ep) & Final-token hidden state, linear head & 0.2860 & {\scriptsize $\pm$0.0066} \\
    \bottomrule
  \end{tabular}
  \caption{\textbf{Per-retriever encoding wins}: joint-prompt encoding: all $R$ retrievers' documents in a single prompt, one forward pass, vs. the per-retriever encoding used elsewhere in this paper. Best fixed retriever is 0.2782; the frozen final-position variant falls \emph{below} this floor. All three variants underperform the per-retriever pipeline (0.3552, Table~\ref{tab:main}).}
  \label{tab:joint-prompt}
\end{table}

These runs chronologically predate the contrastive fine-tuning and comparison-head experiments of Sections~\ref{sec:method} and \ref{sec:results} and were not pursued further once the per-retriever encoding direction showed clearer gains. We report them here as a completed ablation over the encoding architecture itself, rather than omitting them because they came first in time: cramming all retrievers into a single prompt and reading off a single position -- whether via the yes-logit or a linear head on the final hidden state -- does not give the backbone a strong enough per-retriever signal.

\section{What the Scorer's Representations Retain}
\label{sec:appendix-collapse}

The comparison head's gain over the scalar scorer is small (Section~\ref{sec:results-effective}). This appendix measures what the scorer's representations retain, which explains why.

\begin{table}[h]
  \centering
  \small
  \begin{tabular}{@{}lr@{}}
    \toprule
    \textbf{Representation diagnostic} & \textbf{Value} \\
    \midrule
    Mean pairwise cosine similarity & 0.9983 \\
    Share of retriever pairs above 0.99 & 99.6\% \\
    Spearman($\Delta$utility, cosine distance) & 0.085 \\
    \midrule
    Linear probe on 1024-d representations & 0.3059 \\
    Linear probe on the 5 scalar scores & 0.3419 \\
    \bottomrule
  \end{tabular}
  \caption{\textbf{Routing signal concentrates in the score}: the five retrievers' hidden states are highly similar, and a linear probe on them recovers less than the same probe on the five scores.}
  \label{tab:collapse}
\end{table}

\paragraph{Routing signal concentrates in the score} Table~\ref{tab:collapse} measures what the scalar scorer's hidden states retain. The scorer is trained to express its judgment through one number per candidate, and that is where the routing signal ends up. The five retrievers' representations for a query are highly similar (mean pairwise cosine $0.9983$, with $99.6\%$ of pairs above $0.99$), as one would expect of encodings of the same query under the same instruction, and the remaining differences correlate only weakly with which retriever is better (Spearman $0.085$). A linear probe on the full $1024$-dimensional representations recovers $0.3059$ F1, while the same probe on the five scores recovers $0.3419$: what is linearly accessible is already in the score. This says nothing against the scorer, whose score is what routes. It explains why a head reading these representations has limited room, and why, as the next paragraph shows, architecture has been a smaller lever than supervision.

\paragraph{Head architecture} Adding the comparison head on top of these frozen representations reaches $0.3552 \pm 0.0026$ ($+1.4$ over the scalar scorer alone), so a joint, non-linear comparison does recover signal that the linear probe misses. We swept roughly twenty further configurations against the same frozen representations (listed in Appendix~\ref{sec:appendix-sweep}), and none exceeded $\sim$0.353. The only change that moved the number at all was training the head jointly with the backbone (Table~\ref{tab:head-sweep}), and even that tops out at $0.3623$, inside the range we would call noise if it came from a seed change rather than an architecture change. Given Table~\ref{tab:collapse}, this is the expected outcome: most of the routing signal is already in the score, so a more elaborate comparison has little more to find.


\paragraph{An expected effect of scalar supervision} The pattern in Table~\ref{tab:collapse} is what one should expect from supervising a single scalar per retriever: the objective asks for a score, and the model is optimized for that score. It is a property of how the scorer is trained and not a defect of it. The scalar scorer is already content-grounded and utility-supervised, so its large share of the gain (Section~\ref{sec:results-effective}) is evidence for the formulation.

\section{Head Architecture and Objective Sweep}
\label{sec:appendix-sweep}

Appendix~\ref{sec:appendix-collapse} reports that no head configuration in a sweep of roughly twenty exceeds $\sim$0.353 on a frozen backbone. Table~\ref{tab:head-sweep} gives the full breakdown by family. Every row uses a frozen backbone except the last.

\begin{table}[h]
  \centering
  \small
  \setlength{\tabcolsep}{2pt}
  \begin{tabular}{@{}lr@{}}
    \toprule
    \textbf{Family} & \textbf{Range}\\
    \midrule
    Frozen head, metric-learning backbone & $0.3442$--$0.3528$ \\
    Frozen head + yes-logit fusion & $0.3438$--$0.3498$ \\
    Frozen head, alternate backbone (E2E CoT) & $0.3465$--$0.3480$ \\
    Frozen head, alternate backbone (own E2E) & $0.3598$--$0.3599$ \\
    Full end-to-end (backbone + head jointly) & $0.3596$--$0.3623$ \\
    \bottomrule
  \end{tabular}
  \caption{\textbf{The head saturates}: every family of head configuration tried, spanning pooling, depth, dropout, a retriever-identity embedding, loss mode, target temperature, extraction granularity, the source backbone, and fusing the head with the raw scalar score. The two ``alternate backbone'' rows re-cache from a backbone already fine-tuned end-to-end with a head, to test whether a better-trained backbone gives a frozen head more to work with; each reproduces that backbone's own end-to-end number but does not exceed it.}
  \label{tab:head-sweep}
\end{table}

The only family that moves the number materially is training the backbone and head together (last row); every purely architectural or objective change to a frozen head stays within the same $\sim$0.35 band. This is the basis for treating the head's contribution as bounded by what the frozen representations retain beyond the score (Table~\ref{tab:collapse}), not by its own capacity.

\section{Retrieval Metrics}
\label{sec:appendix-retrieval}

Section~\ref{sec:results-effective} reports the router's effect on answer quality, the quantity it is fit to. It also improves the intermediate metrics prior retrieval work optimizes, which we give here in full.

\begin{figure}[ht]
  \centering
  \includegraphics{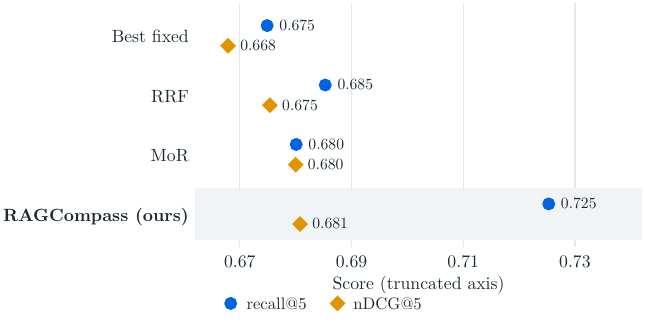}
  \caption{\textbf{Recall improves clearly}: retrieval recall@5 and nDCG@5 by method. Our recall gain is clear; the nDCG gain is small and roughly matched by the fusion baselines.}
  \label{fig:retrieval-metrics}
\end{figure}

The routed choice reaches recall@5 $0.7253 \pm 0.0031$ and nDCG@5 $0.6808 \pm 0.0068$, against $0.6749$ and $0.6679$ for the best single fixed retriever (HippoRAG2 and SPLADE-v3 respectively -- the two metrics do not pick the same one). That is $+5.0$ points of recall, ahead of both fusion baselines ($0.6853$ RRF, $0.6801$ MoR), and $+1.3$ of nDCG, roughly level with them ($0.6754$, $0.6800$; Figure~\ref{fig:retrieval-metrics}). Answer quality improves by more than retrieval quality does, as one expects of a router supervised on the answer rather than on relevance.

\section{Retriever Agreement and Routing Decisions}
\label{sec:appendix-agreement}

Two further observations complement Section~\ref{sec:results-filter}: whether agreement between retrievers is a usable routing signal on its own, and how the router's choices are distributed.

\paragraph{Majority vote} Taking the modal answer across the five retrievers' independently generated answers, with no learned router at all, underperforms \emph{both} the best fixed retriever and picking one at random ($0.2009$ against $0.2782$ and $0.2473$): agreement between retrievers is anti-correlated with correctness here, not a usable signal on its own.

\paragraph{Routing decisions} On the reported population, the router's choice distribution tracks the oracle's within about $6$ points per retriever, the one systematic bias being that BM25 is under-selected.

\section{Generalization Results, Visualized}
\label{sec:appendix-transfer}

\begin{figure}[ht]
  \begin{minipage}[t]{0.485\textwidth}
    \centering
    \includegraphics{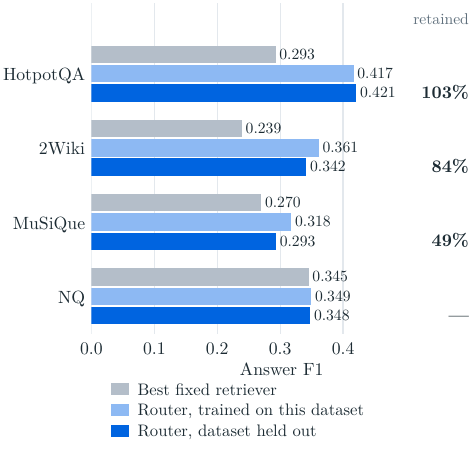}
    \caption{\textbf{Held-out tracks in-distribution}: leave-one-dataset-out: the router trained without a dataset (``held out'') tracks the identical router trained with it (``in-distribution'') far more closely than either tracks the best fixed retriever, on every dataset where routing has headroom to begin with.}
    \label{fig:corpus-transfer}
  \end{minipage}\hfill
  \begin{minipage}[t]{0.485\textwidth}
    \centering
    \includegraphics{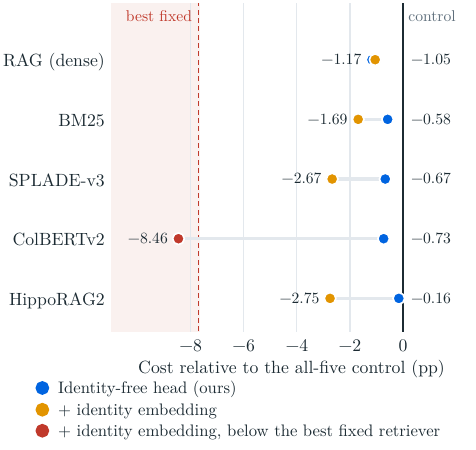}
    \caption{\textbf{Identity-free stays stable}: cost of leaving one retriever out of training, relative to the all-five control. Each row pairs the two heads on the same held-out retriever, so the connector is the penalty the identity embedding pays. The identity-free head (blue) stays within $1.2$pp on every fold; the identity embedding (orange) is far more volatile, and its ColBERTv2 fold (red) crosses below the best-fixed floor -- worse than not routing.}
    \label{fig:mechanism-transfer}
  \end{minipage}
\end{figure}

Figures~\ref{fig:corpus-transfer} and Figure~\ref{fig:mechanism-transfer} show the visualization of the transfer results reported in Section~\ref{sec:results-generalization}. Both restate Tables~\ref{tab:corpus-transfer} and~\ref{tab:mechanism-transfer} rather than adding evidence, and are kept for readers who prefer the visual form.

\section{Unseen Retrieval Depth}
\label{sec:appendix-depth}

\begin{figure}[ht]
  \centering
  \includegraphics{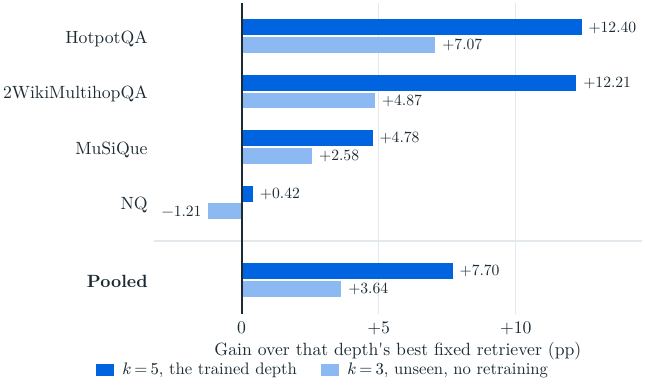}
  \caption{\textbf{Generalizes across depth}: gain over the best fixed retriever at the trained depth ($k{=}5$) and at an unseen shallower depth ($k{=}3$), with no retraining. Each bar is measured against its own depth's best fixed retriever; those are different systems, so the pairs share an axis but not a floor.}
  \label{fig:depth-transfer}
\end{figure}

Retrieval depth is fixed at $k{=}5$ throughout training. Re-answering the same test queries from only the top $3$ passages -- same reader, same prompt, same retrieved ranking, and no retraining of the router -- the router reaches $0.3343$ against $0.2979$ for the best fixed retriever at that depth ($+3.64$pp). The absolute gain is roughly half the $+7.70$pp at $k{=}5$, but so is the headroom: the oracle falls from $0.7508$ to $0.5645$, because fewer questions are answerable from three passages by \emph{any} retriever. As a share of what is reachable the router captures $13.7\%$ against $16.3\%$ at its trained depth, and the per-dataset ordering is unchanged (Figure~\ref{fig:depth-transfer}), with NQ the smallest at both depths ($+0.42$pp at $k{=}5$, negative at $k{=}3$). Each depth is scored against its own best fixed retriever, whose identity changes (SPLADE-v3 to HippoRAG2), and the shift is downward only (Limitations).

\section{Training-Set-Size Ablation}
\label{sec:appendix-datasize}

The leave-one-dataset-out protocol (Section~\ref{sec:results}) necessarily trains on three datasets instead of four. Table~\ref{tab:datasize-ablation} isolates whether that reduction in training data, by itself, costs anything -- independent of the corpus and query-distribution shift leave-one-dataset-out also introduces -- by training on three datasets and evaluating on those same three, rather than on the held-out fourth.

\begin{table}[h]
  \centering
  \small
  \setlength{\tabcolsep}{3pt}
  \begin{tabular}{@{}lrrr@{}}
    \toprule
    \textbf{Trained without} & \textbf{Router} & \textbf{Fixed} & \textbf{Gain} \\
    \midrule
    NQ & 0.3542 {\scriptsize $\pm$0.0026} & 0.2562 & $+9.80$pp \\
    MuSiQue & 0.3621 {\scriptsize $\pm$0.0013} & 0.2820 & $+8.01$pp \\
    HotpotQA & 0.3389 {\scriptsize $\pm$0.0028} & 0.2747 & $+6.42$pp \\
    2Wiki & 0.3509 {\scriptsize $\pm$0.0024} & 0.3002 & $+5.07$pp \\
    \bottomrule
  \end{tabular}
  \caption{\textbf{Not a data-size artifact}: training on three datasets, evaluated on those same three (not the fourth); ``Fixed'' is the best fixed retriever. Against the four-dataset control, $0.3552 \pm 0.0026$, dropping a quarter of the training pool costs little: the router is not starved for training data at this scale. This isolates data quantity from the corpus and query-distribution shift that Table~\ref{tab:corpus-transfer} tests.}
  \label{tab:datasize-ablation}
\end{table}

None of these four numbers is a transfer result, since each is evaluated on data the training mix already included. The table rules out a confound in Table~\ref{tab:corpus-transfer}'s transfer numbers: training on a quarter less data, by itself, costs little, so the held-out results reflect the corpus and query shift rather than reduced training data.

\section{Baseline Implementation Details}
\label{sec:appendix-r3ag}

All three learned query-only baselines are our reimplementations, following experimental details in the corresponding papers. They run on the same filtered population, splits and seeds as our router. Each differs from its original system, because our retriever pool, reader and compute budget differ from the ones the methods were designed for. Table~\ref{tab:baseline-diff} summarizes the main differences.

\begin{table}[!t]
  \centering
  \footnotesize
  \setlength{\tabcolsep}{3pt}
  \begin{tabular}{@{}>{\raggedright\arraybackslash}p{0.25\linewidth}>{\raggedright\arraybackslash}p{0.34\linewidth}>{\raggedright\arraybackslash}p{0.34\linewidth}@{}}
    \toprule
    \textbf{Component} & \textbf{Original} & \textbf{Ours} \\
    \midrule
    \multicolumn{3}{@{}l}{\textbf{R\textsuperscript{3}AG}} \\
    Retrieval-quality judge & $80$B model & Qwen2.5-7B-Instruct \\[2pt]
    Candidate pool & includes a null (no-retrieval) retriever & five retrievers, no null retriever \\[2pt]
    Batch size & effective $512$ & $16$ \\[2pt]
    Fusion-stage settings & not specified & learning rate $10^{-4}$, $10$ epochs, chosen on validation \\[2pt]
    Reader exact match & about $36\%$ & $4$--$10\%$ (long answers) \\
    \addlinespace
    \multicolumn{3}{@{}l}{\textbf{LTRR}} \\
    Ranking model & pairwise XGBoost (learned LTR router) & logistic regression, random forest or gradient boosting, chosen on validation \\[2pt]
    Features
        & query embedding, length, type; embedding-similarity features of retrieved documents
        & regular-expression query features and per-retriever retrieval-score statistics \\[2pt]
    Formulation
        & pairwise ranking of six retrievers plus a null option, trained on utility gain
        & five-way classifier; shared (query, retriever) scorer for the unseen-mechanism test \\
    \addlinespace
    \multicolumn{3}{@{}l}{\textbf{R.RETRIEVER}} \\
    Experts & domain-adapted instances of one embedding model & five structurally different retrievers \\[2pt]
    Centroid membership & an expert's own domain queries & training queries where that retriever's F1 is highest \\[2pt]
    Query encoder & the experts' shared base encoder & Qwen3-Embedding-0.6B \\
    \addlinespace
    \multicolumn{3}{@{}l}{\textit{MoR}} \\
    Retriever weights & per query, zero-shot, from embedding-geometry signals & one weight per retriever, fit on validation utility \\[2pt]
    Retriever pool & each retriever at four index granularities & five retrievers, one index each \\
    \bottomrule
  \end{tabular}
  \caption{\textbf{How our reimplementations differ from the originals}: main differences only; details in the text.}
  \label{tab:baseline-diff}
\end{table}

\paragraph{R\textsuperscript{3}AG} We follow \citet{zhao_r3ag_2026} wherever the paper specifies a choice: a single trainable identity token per retriever, never a description or its outputs (their Appendix~B); two Qwen3-Embedding-0.6B encoders with only the word embeddings and the last two layers trainable, learning rate $2\times10^{-6}$, $10$ epochs (their A.5); a top-$2$ multi-positive contrastive loss with $\tau{=}0.1$; utility $\mathrm{EM} + 0.2\,\mathrm{F1} + 0.2\,\sigma(R)$, with $\sigma$ the retriever's mean training EM; attention fusion of each candidate's two capability vectors conditioned on the query; cosine routing; binary cross-entropy on the sigmoid of that cosine plus a squared-$\ell_2$ regularizer. We depart from it in five places. (i) The pool excludes their null retriever, so R\textsuperscript{3}AG is scored on exactly the population of Table~\ref{tab:main}: adding an action changes which queries pass the utility-gap filter. (ii) The retrieval-quality judge is Qwen2.5-7B-Instruct rather than an $80$B model, using their five-dimension rubric; its averaged integer scores tie at the top-$2$ boundary on about $64\%$ of training queries, and we break ties at random. (iii) We train with batch size $16$ rather than an effective $512$, which on our $5{,}009$ training queries would allow about $100$ optimizer steps in total. (iv) The fusion stage's learning rate ($10^{-4}$), epochs ($10$) and regularizer weight are not given; we select the stage-1 learning rate, regularizer weight and $\gamma$ from a five-point grid on validation F1, which returns the paper's learning rate and $\gamma$ with a regularizer weight of $0.1$, and pick the fusion epoch on validation. (v) Our reader's answers are long, so EM is $4$--$10\%$ against roughly $36\%$ in their setting; the EM term of the utility is mostly zero and the per-retriever prior $\sigma$ often decides the second positive. On validation, no configuration in the grid beats the best fixed retriever; the one selected routes nearly every query to it.

\paragraph{LTRR-style ranker} We keep LTRR's formulation, a learned ranker over retrievers supervised by downstream utility and fed query and retrieval-confidence features. The original trains pairwise XGBoost on normalized utility gain, using a PCA-reduced query embedding, query length and type, and six embedding-similarity features of the retrieved documents (e.g., MaxSim, VarSim, Moran). The five-way classifier and the shared per-retriever scorer each fit several model families (logistic regression, random forest and gradient boosting, plus two regression targets for the shared scorer); the family is chosen per run on validation F1 and refit on training plus validation for the test number. Its in-distribution result depends on the gradient-boosting fit more than on the formulation: with the model family fixed to gradient boosting and seed $42$, the same code and data give $0.2903$ under scikit-learn 1.6 and 1.7 but $0.2782$ and $0.2785$ under 1.5 and 1.9, while logistic regression is identical across versions. We report three seeds under a single library version (1.9) with validation selection. This sensitivity, up to $1.2$pp, is larger than the ranker's margin over the best fixed retriever.

\paragraph{R.RETRIEVER} \citet{lee2025routerretrieverroutingmixtureexpert}'s routing has no learned parameters: each expert's ``pilot embedding'' is the centroid of a shared query encoder's embeddings for that expert's own training queries, and a query routes to the nearest centroid by cosine similarity. Their experts are domain-adapted instances of one embedding architecture, so ``that expert's own training queries'' is domain membership, independent of any other expert. We route across five structurally different retrievers instead, with no such shared notion of a retriever's ``own'' queries, so we substitute the closest available analogue: with Qwen3-Embedding-0.6B as the shared encoder, retriever $i$'s centroid is built from the training queries where retriever $i$'s own F1 is highest -- a per-retriever, non-comparative signal, unlike the argmax label every other baseline in this paper uses. The one hyperparameter, the fraction of training queries contributing to each centroid, is chosen per run from $\{0.1, 0.2, 0.35, 0.5, 1.0\}$ on validation F1 ($1.0$ is a naive full-mean centroid, ignoring F1 entirely). Being training-free, it has nothing to overfit to the training corpus, but that is not the same as being accurate: it is below the best fixed retriever in distribution ($-1.74$pp) and on every leave-one-dataset-out fold ($-2.07$ to $-8.01$pp, worst on MuSiQue). We do not report a leave-one-retriever-out number: retriever $i$'s centroid needs retriever $i$'s own F1 outcomes to be built at all, comparative or not, which is exactly what that protocol withholds, so its score for a held-out retriever cannot be constructed without reusing information the protocol is meant to deny it, the same wall the plain LTRR classifier hits.

\paragraph{Fusion baselines} RRF uses the standard constant $k{=}60$~\citep{cormack_reciprocal_2009} and no training. The original MoR~\citep{kalra_mor_2025} is zero-shot: it weights each retriever per query using embedding-geometry signals (distance to corpus cluster centroids, Moran coefficient) with fixed mixing coefficients, and also fuses over several index granularities. Ours instead fits one weight per retriever on validation utility and applies it to every query, with no per-query signals or multi-granularity indices.

\paragraph{Implications for the comparison} The differences above go in one direction: our versions use smaller models and less compute than the originals, and two of the methods are applied to a pool of structurally different retrievers that they were not designed for. We therefore expect our numbers to understate what the original systems achieve in their own settings, and they should not be compared with the numbers in the corresponding papers. Our MoR variant differs in kind and not in scale: it is not zero-shot and does not adapt to the query. R\textsuperscript{3}AG's near-collapse onto a single retriever (Section~\ref{sec:results-effective}) may partly reflect these conditions, in particular the weaker judge and the near-zero exact-match term, and not only the method.

\section{Per-Dataset Breakdown}
\label{sec:appendix-perdataset}

\begin{table}[!t]
  \centering
  \small
  \setlength{\tabcolsep}{3pt}
  \begin{tabular}{@{}lrrrr@{}}
    \toprule
     & \textbf{HotpotQA} & \textbf{2Wiki} & \textbf{MuSiQue} & \textbf{NQ} \\
    \midrule
    $n$ & 724 & 896 & 1195 & 920 \\
    Routed & 0.4168 & 0.3614 & 0.3178 & 0.3495 \\
     & {\scriptsize $\pm$0.0017} & {\scriptsize $\pm$0.0020} & {\scriptsize $\pm$0.0042} & {\scriptsize $\pm$0.0058} \\
    Best fixed & 0.2928 & 0.2393 & 0.2700 & 0.3453 \\
    Oracle & 0.7897 & 0.7987 & 0.7401 & 0.6873 \\
    Gain (pp) & $+12.40$ & $+12.21$ & $+4.78$ & $+0.42$ \\
    Captured & 25.0\% & 21.8\% & 10.2\% & 1.2\% \\
    \bottomrule
  \end{tabular}
  \caption{\textbf{Gain tracks headroom}: per-dataset breakdown of the reported system. ``Captured'' is the headroom captured: the gain as a share of the room between that dataset's best fixed retriever and its oracle. Each row is scored against \emph{its own} dataset's best fixed retriever -- the same convention as Table~\ref{tab:corpus-transfer} -- which is a stricter floor than the single global best fixed retriever (SPLADE-v3, $0.2782$) that Table~\ref{tab:main} reports against; pooled against these per-dataset floors the gain is $+7.0$pp rather than $+7.70$pp. Routed: mean $\pm$ std over three seeds}
  \label{tab:per-dataset}
\end{table}

The pooled result is not evenly earned (Table~\ref{tab:per-dataset}), and NQ, a quarter of the test set, contributes essentially nothing. The gain follows available headroom, the room between a dataset's best fixed retriever and its oracle, and not hop count. NQ's headroom is the smallest of the four ($0.342$), and SPLADE-v3 alone already reaches $0.3453$ there, so there is little for a router to win; on HotpotQA, where $0.497$ separates floor from oracle, routing captures a quarter of it. The transfer study above reproduces the same pattern, with retention likewise tracking headroom rather than dataset identity (Table~\ref{tab:corpus-transfer}). Routing helps where there is something to choose between, and single-hop NQ is a case where there largely is not.

\section{An LLM as Router, and the Cost of Routing}
\label{sec:appendix-cost}

\paragraph{An LLM as router} Prompting Llama-3.1-8B-Instruct (roughly $13\times$ our router's parameters) with the query and all five evidence sets and asking it to choose zero-shot gains $0.20$pp over the best fixed retriever; asked instead to choose among the five already-generated answers, it gains $3.20$pp. Our $0.6$B router, fine-tuned against downstream utility on the same queries, gains $7.70$pp. The evidence-judging variant is also strongly position-biased, picking whichever evidence set comes first on $51.8\%$ of queries against a $20\%$ unbiased rate, and it costs more: $0.358$s/query against $0.222$s for our router, which generates nothing. We attribute the gap to the training signal rather than to model capacity, and it is not bought with compute.

\paragraph{Cost} Reading the evidence means running all $R$ retrievers before deciding, which a single-retriever system does not pay. That cost is small compared to the cost of the reader: the decision itself takes $0.222$s/query (corresponding to five $0.6$B forward passes) against $5.687$s for one retriever's full retrieve-and-answer pipeline, an overhead of $3.9\%$, and the $R$ retrievals are independent and parallelizable. Generation, the expensive step, happens once either way. 

We estimate the reported experiments at roughly $700$ GPU-hours on NVIDIA L40S GPUs: about $475$ for running the five retrievers and the reader over all $60{,}000$ queries to obtain utility labels, and about $200$ for router training, generalization folds, ablations and baselines; index construction and preliminary experiments are excluded.